\documentclass[10pt,twocolumn,letterpaper]{article}

\usepackage{btas}
\usepackage{times}
\usepackage{epsfig}
\usepackage{graphicx}
\usepackage{amsmath}
\usepackage{amssymb}

\graphicspath{{DeepPyramidFigs/}}

\btasfinalcopy 

\ifbtasfinal\fi
\begin{document}

\title{A Deep Pyramid Deformable Part Model for Face Detection}

\author{Rajeev Ranjan, Vishal M. Patel, Rama Chellappa\\
Center for Automation Research\\
University of Maryland, College Park, MD 20742\\
{\tt\small\{rranjan1, pvishalm, rama\}@umiacs.umd.edu}
}

\maketitle
\thispagestyle{empty}

\begin{abstract}
We present a face detection algorithm based on Deformable Part Models and deep pyramidal features. The proposed method called DP2MFD is able to detect faces of various sizes and poses in unconstrained conditions. It reduces the gap in training and testing of DPM on deep features by adding a normalization layer to the deep convolutional neural network (CNN).  Extensive experiments on four publicly available unconstrained face detection datasets show that our method is able to capture the meaningful structure of faces and performs significantly better than many competitive face detection algorithms.
\end{abstract}

\section{Introduction}
Face detection is a challenging problem that has been actively researched for over two decades \cite{face_survey}, \cite{MSR_FD_TR_2012}.   Current methods work well on images that are captured under user controlled conditions.  However, their performance degrades significantly on images that
have cluttered backgrounds and have large variations in face viewpoint, expression, skin color, occlusions and cosmetics.

The seminal work of Viola and Jones \cite{Viola_Jones} has made face detection feasible in real world applications. They use cascaded classifiers on Haar-like features to detect faces. The cascade structure has been a subject of extensive research since then. Cascade detectors work well on frontal faces, however, sometimes they fail to detect profile or partially occluded faces. A recently developed joint cascade-based  method \cite{JointCascade_LI_ECCV2014} yields improved detection performance by incorporating a face alignment step in the cascade structure. Headhunter \cite{HeadHunter_Mathias_ECCV2014} uses rigid templates along similar lines.  The method based on Aggregate Channel Features (ACF) \cite{ACF_multiscale_Yang_IJCB2014} deploys a cascade of channel features while Pixel Intensity Comparisons Organized (Pico) \cite{PICO_Markus_CoRR2014} uses a cascade of rejectors for improved face detection.

Most of the recent face detectors are based on the Deformable Parts Model (DPM) structure \cite{DPM_PAMI2010} where a face is defined as a collection of parts. These parts are trained side-by-side with the face using a spring-like constraint. They are fine-tuned to work efficiently with the HOG \cite{HOG} features.  A unified approach for face detection, pose estimation and landmark localization using the DPM framework was recently proposed in \cite{AFW_dataset_CVPR2012}.  This approach defined a ``part" at each facial landmark and used mixture of tree-structured models resilient to viewpoint changes. A properly trained simple DPM is shown to yield significant improvement for face detection in \cite{HeadHunter_Mathias_ECCV2014}.

The key challenge in unconstrained face detection is that features like Haar wavelets and HOG do not capture the salient facial information at different poses and illumination conditions. The limitation is more due to the features used than the classifiers. However, with recent advances in deep learning techniques and the availability of GPUs, it is becoming possible to use deep Convolutional Neural Networks (CNN) for feature extraction. In has been shown in \cite{NIPS2012_4824} that a deep CNN pretrained with a large generic dataset such as Imagenet \cite{deng2009imagenet}, can be used as a meaningful feature extractor. The deep features thus obtained have been used extensively for object detection.  For instance, Regions with CNN (R-CNN) \cite{girshick14CVPR} computes regions-based deep features and attains state-of-art on the Imagenet challenge. Methods like Overfeat \cite{Overfeat} and Densenet \cite{Densent} adopt a sliding window approach to detect objects from the $pool_{5}$ features. Deep Pyramid \cite{DPM_r_CNN} and  Spatial Pyramid \cite{DBLP:journals/corr/HeZR014} remove the fixed-scale input dependency from deep CNNs which makes them attractive to be integrated with DPMs. Although, a lot of research on deep learning has focused on object detection and classification, very few have used deep features for face detection which is equally challenging because of high variations in pose, ethnicity, occlusions, etc. It was shown in \cite{multiviewFD_RCNN}  that deep CNN features fine-tuned on faces are informative enough for face detection, and hence do not require an SVM classifier. They detect faces based on the heat map score obtained directly from the fifth convolutional layer. Although they report competitive results, detection performance for faces of various sizes and occlusions needs improvement.

\begin{figure*}[htp!]
 \centering
 \includegraphics[width=16cm]{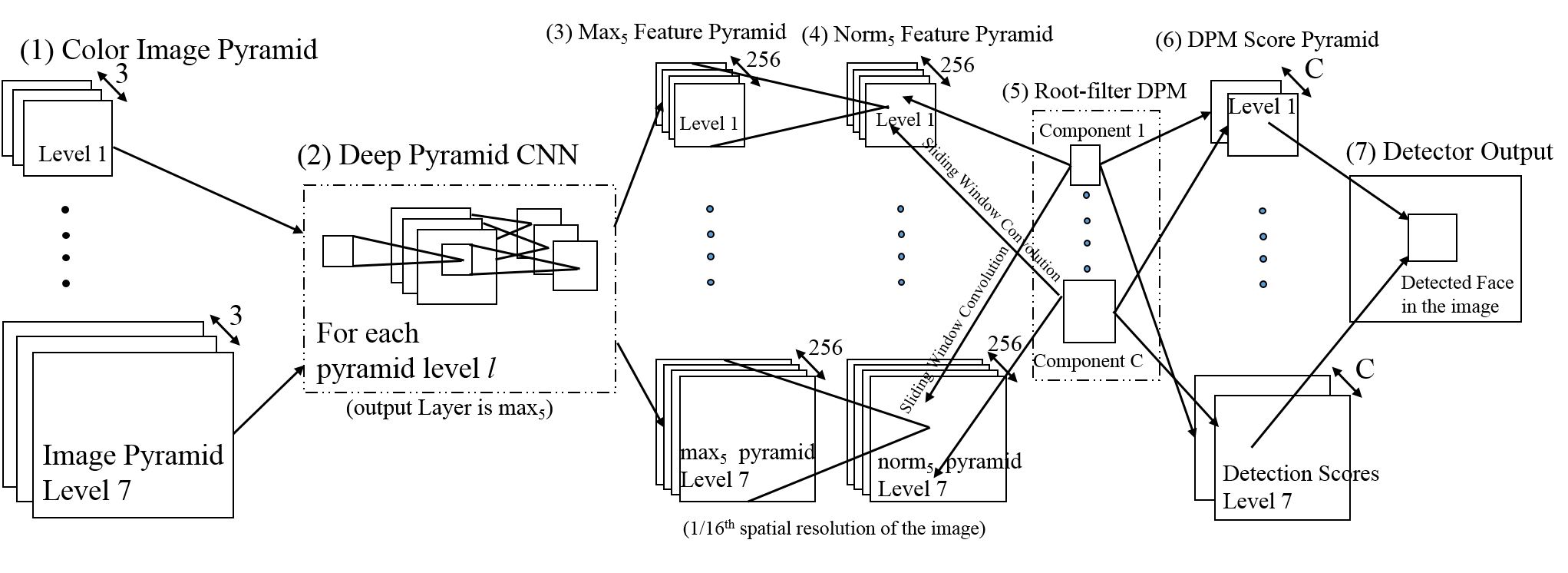}\\
 \caption{Overview of our approach. (1) An image pyramid is built from a color input image with level 1 being the lowest size. (2) Each  pyramid level is forward propagated through a deep pyramid CNN \cite{DPM_r_CNN} that ends at max variant of convolutional layer 5 ($max_{5}$). (3) The result is a pyramid of $max_{5}$ feature maps, each at 1/16th the spatial resolution of its corresponding image pyramid level. (4) Each $max_{5}$ level features is normalized using $z$-score to form $norm_{5}$ feature pyramid. (5) Each $norm_{5}$ feature level gets convoluted with every root-filter of a C-component DPM to generate a pyramid of DPM score (6). The detector outputs a bounding box for face location (7) in the image after non-maximum suppression and bounding box regression.}
\label{fig:DP2MFD}
\end{figure*}

In this paper, we propose a face detector which detects faces at multiple scales, poses and occlusion by efficiently integrating deep pyramid features \cite{DPM_r_CNN} with DPMs.  This paper makes the following contributions:
\begin{enumerate}
\item We propose a novel method for training DPM for faces using deep pyramidal features.
\item We propose adding a normalization layer to the deep CNN to reduce the bias in face sizes.
\item We achieve new state-of-the-art detection performances on four challenging face detection datasets.
\end{enumerate}

This paper is organized as follows. Section \ref{method} describes our proposed face detector in detail. Section \ref{results} provides the detection results on four challenging datasets.  Finally, Section \ref{conclusion} concludes the paper with a brief summary and discussion.

\begin{figure*}[htp!]
 \centering
 \includegraphics[width=13cm]{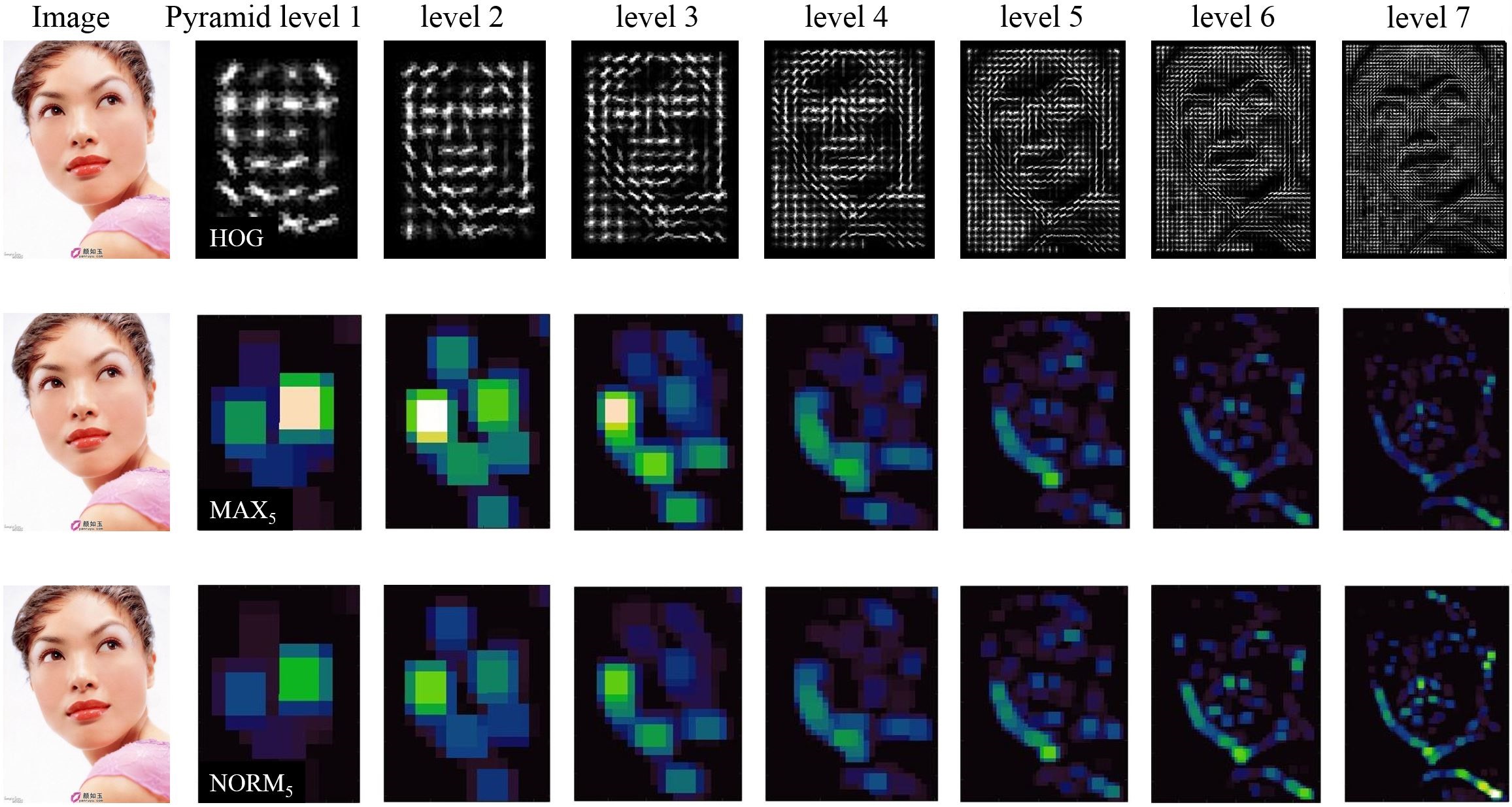}\\
 \caption{Comparison between HOG, $max_{5}$ and $norm_{5}$ feature pyramids. In contrast to $max_{5}$ features which are scale selective, $norm_{5}$ features have almost uniform activation intensities across all the levels. }
\label{fig:DP2MFDFeatures}
\end{figure*}

\section{Face Detection with Deep Pyramid DPM} \label{method}
 Our proposed face detector, called Deep Pyramid Deformable Parts Model for Face Detection (DP2MFD), consists of two modules. The first one generates a seven level normalized deep feature pyramid for any input image of arbitrary size. Fixed-length features from each location in the pyramid are extracted using the sliding window approach. The second module is a linear SVM which takes these features as input to classify each location as face or non-face, based on their scores. In this section, we provide the design details of our face detector and describe its training and testing processes.

 \subsection{DPM Compatible Deep Feature Pyramid}
 We build our model using the feature pyramid network implementation provided in \cite{DPM_r_CNN}. It takes an input image of variable size and constructs an image pyramid with seven levels. Each level is embedded in the upper left corner of a large ($1713 \times 1713$ pixels) image and maintains a scale factor of $\sqrt{2}$ with its next lower level in the hierarchy. Using this image pyramid, the network generates a pyramid of 256 feature maps at the fifth convolution layer ($conv_{5}$). A $3 \times 3$ max filter is applied to the feature pyramid at a stride of one to obtain the $max_{5}$ layer which essentially incorporates the $conv_{5}$ ``parts" information. Hence, it suffices to train a root-only DPM on the $max_{5}$ feature maps without explicitly training on DPM parts.  A cell at location $(j,k)$ in the $max_{5}$ layer corresponds to the pixel $(16j,16k)$ in the input image, with a highly overlapping receptive field of size $163 \times 163$ pixels. Despite having a large receptive field , the features are well localized to be effective for sliding window detectors.

 It has been suggested in \cite{DPM_r_CNN} that deep feature pyramids can be used as a replacement for HOG Pyramid in DPM implementation.  However, this is not entirely obvious as deep features are different than HOG features in many aspects. Firstly, the deep features from $max_{5}$ layer have a receptive field of size $163 \times 163$ pixels, unlike HOG where the receptive region is localized to a bin of $8 \times 8$ pixels. As a result, $max_{5}$ features at face locations in the test images would be substantially different from that of a cropped face. This prohibits us from using the deep features of cropped faces as positive training samples, which is usually the first step in training HOG-based DPM. Hence, we take a different approach of collecting positive and negative training samples from the deep feature pyramid itself.   This procedure is described in detail in subsection \ref{training}.

 Secondly, the deep pyramid features lack the normalization attribute associated with HOG. The feature activations vary widely in magnitude across the seven pyramid levels as shown in Figure~\ref{fig:DP2MFDFeatures}. Typically, the activation magnitude for a face region decreases with the size of pyramid level. As a result, a large face detected by a fixed-size sliding window at a lower pyramid level will have a high detection score compared to a small face getting detected at a higher pyramid level. In order to reduce this bias to face size, we apply a $z$-score normalization step on the $max_{5}$ features at each level. For a 256-dimensional feature vector $x_{i,j,k}$ at the pyramid level $i$ and location $(j, k)$, the normalized feature $\hat{x}_{i,j,k}$ is computed as:
 \begin{equation}
 \hat{x}_{i,j,k} = \frac{x_{i,j,k} - \mu_{i}}{\sigma_{i}},
 \end{equation}
where $\mu_{i}$ is the mean feature vector, and $\sigma_{i}$ is the standard deviation for the pyramid level $i$. We refer to the normalized $max_{5}$ features as "$norm_{5}$". A root-only DPM is trained on the $norm_{5}$ feature pyramid using a linear SVM.  Figure~\ref{fig:DP2MFD} shows the complete overview of our model.

\subsection{Testing}
At test time, each image is fed to the model described above to obtain the $norm_{5}$ feature pyramid. They are convolved with the fixed size root-filters for each component of DPM in a sliding window fashion, to generate a detection score at every location of the pyramid. Locations having scores above a certain threshold are mapped to their corresponding regions in the image. These regions undergo a greedy non-maximum suppression to prune low scoring detection regions with Intersection-Over-Union (IOU) overlap above 0.3. In order to localize the face as accurately as possible, the selected boxes undergo bounding box regression. Owing to the subsampling factor of 16 between the input image and $norm_{5}$ layer, the total number of sliding windows account to approximately 25k compared to approximately 250k for the HOG pyramid, which reduces the effective test-time.

\subsection{Training} \label{training}
For training, both positive and negative faces are sampled directly from the $norm_{5}$ feature pyramid. The dimensions of root filters for DPM are decided by the aspect ratio distribution for faces in the dataset. The root-filter sizes are scaled down by a factor of 8 to match the face size in the feature pyramid. Since, a given training face maps its bounding box at each pyramid level, we choose the optimal level $l$  for the corresponding positive sample by minimizing the sum of absolute difference between the dimensions of bounding box and the root filter at each level. For a root-filter of dimension $(h, w)$ and bounding box dimension of $(b_{i}^{y},b_{i}^{x})$ for the pyramid level $i$, $l$  is given by
\begin{equation}
l = \arg\min_{i}{|{b_{i}^{y} - h}| + |{b_{i}^{x} - w}|}.
\end{equation}
The ground truth bounding box at level $l$ is then resized to fit the DPM root-filter dimensions. We finally extract the "$norm_{5}$" feature of dimension $h \times w \times 256$ from the shifted ground truth position in the level $l$ as a positive sample for training.

The negative samples are collected by randomly choosing root-filter sized boxes from the normalized feature pyramid. Only those boxes having IOU less than 0.3 with the ground truth face at the particular level are considered as negative samples for training.

Once the training features are extracted, we optimize a linear SVM for each component of the root-only DPM. Since the training data is large to fit in the memory, we adopt the standard hard negative mining method \cite{Sung95examplebased,DPM_PAMI2010} to train the SVM. We also train a bounding box regressor to localize the detected face accurately. The procedure is similar to the bounding box regression used in R-CNN  \cite{girshick14CVPR} , the only difference being our bounding box regressor is trained on the $norm_{5}$ features.

\section{Experimental Results} \label{results}
We evaluated the proposed deep pyramid DPM face detection method on four challenging face detection datasets - Annotated Face in-the-Wild (AFW) \cite{AFW_dataset_CVPR2012}, Face Detection Dataset and Benchmark (FDDB) \cite{fddbTech}, Multi-Attribute Labelled Faces (MALF) \cite{MALF} and the IARPA Janus Benchmark A (IJB-A) \cite{JanusFR_CVPR2015}, \cite{JanusFD_ICB2015} dataset.  We train our detector on the FDDB images using Caffe \cite{jia2014caffe} for both 1-component (DP2MFD-1c) and 2-components (DP2MFD-2c) DPM. The FDDB dataset was evaluated using the 10-fold cross-validation approach. For evaluating the AFW and the MALF datasets, images from all the 10 splits of the FDDB dataset were used as training samples.

\subsection{AFW Dataset Results}
The AFW dataset \cite{AFW_dataset_CVPR2012} contains 205 images with 468 faces collected from Flickr.   Images in this dataset contain cluttered backgrounds with large variations in both face viewpoint and appearance.

\begin{figure}[htp!]
 \centering
  \includegraphics[width=6.5cm]{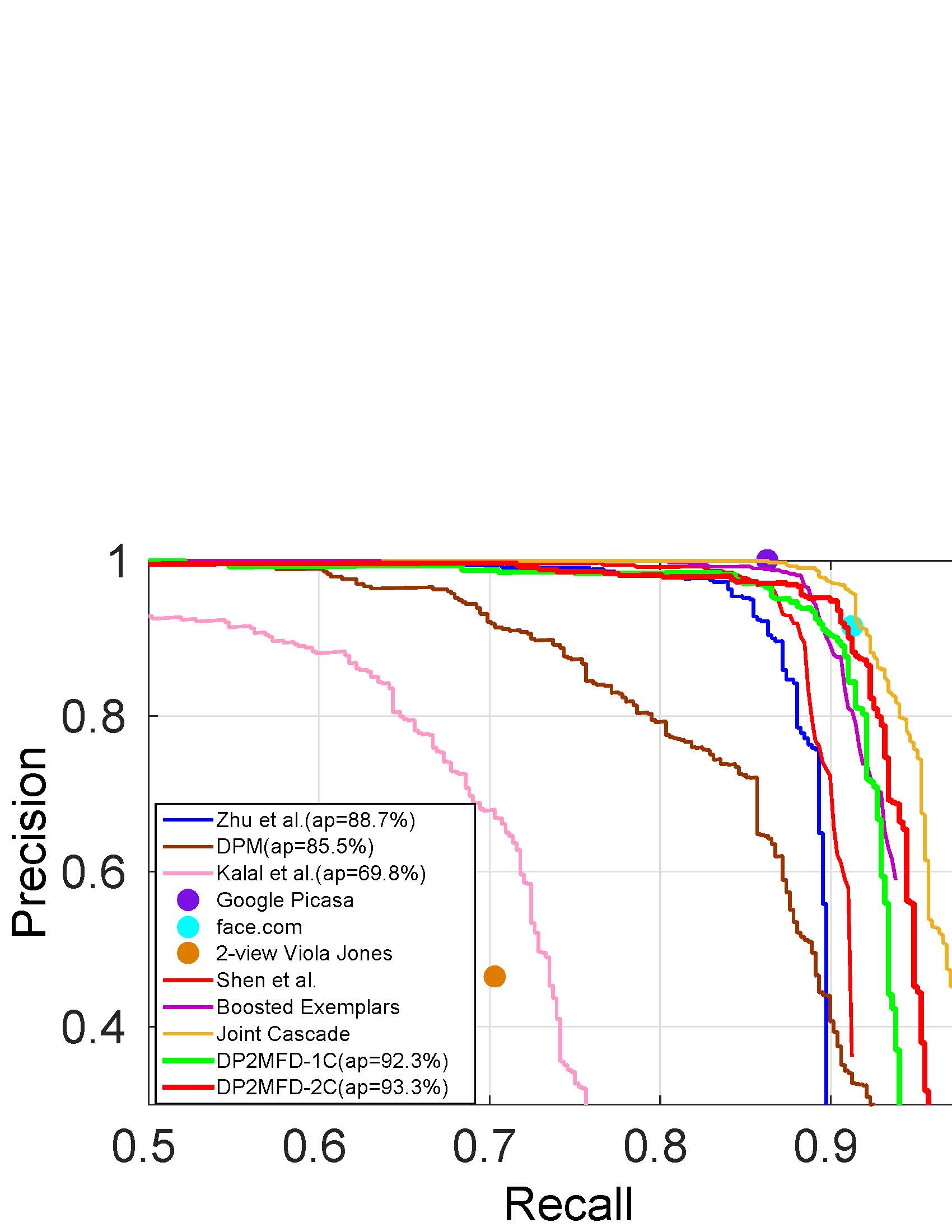}\\
 \caption{Performance evaluation on the AFW dataset.}
\label{fig:AFW_results}
\end{figure}

The precision-recall curves \footnote{The results of the methods other than our DP2MFD methods compared in Figure~\ref{fig:AFW_results} were provided by the authors of \cite{AFW_dataset_CVPR2012}, \cite{JointCascade_LI_ECCV2014} and \cite{BoostedExemplar_Li_CVPR2014}.} of different academic as well as commercial methods on the AFW dataset are shown in Figure~\ref{fig:AFW_results}.  Some of the academic face detection methods compared in Figure~\ref{fig:AFW_results} include OpenCV implementations of the 2-view Viola-Jones algorithm, DPM \cite{DPM_PAMI2010}, mixture of trees (Zhu et al.) \cite{AFW_dataset_CVPR2012}, boosted multi-view face detector (Kalal et al.) \cite{Kalal_BMVC2008}, boosted exemplar \cite{BoostedExemplar_Li_CVPR2014} and the joint cascade methods \cite{JointCascade_LI_ECCV2014}.   As can be seen from this figure, our method outperforms most of the academic detectors and performs comparably to a recently introduced joint cascade-based method  \cite{JointCascade_LI_ECCV2014} and the best commercial face detector Google Picassa.
Note that the joint cascade-based method \cite{JointCascade_LI_ECCV2014} uses face alignment to make the detection better and trains the model on 20,000 images.  In contrast, we do not use any alignment procedure in our detection algorithm and train on only 2,500 images.

\subsection{FDDB Dataset Results}
The FDDB dataset \cite{fddbTech} is the most widely used benchmark for unconstrained face detection.  It consists of 2,845 images containing a total of 5,171 faces collected from news articles on the Yahoo website.  All images were manually localized for generating the ground truth.  The FDDB dataset has two evaluation protocols - discrete and continuous which essentially correspond to coarse match and precise match between the detection and the ground truth, respectively.

\begin{figure*}[htp!]
 \centering
\includegraphics[width=4.2cm]{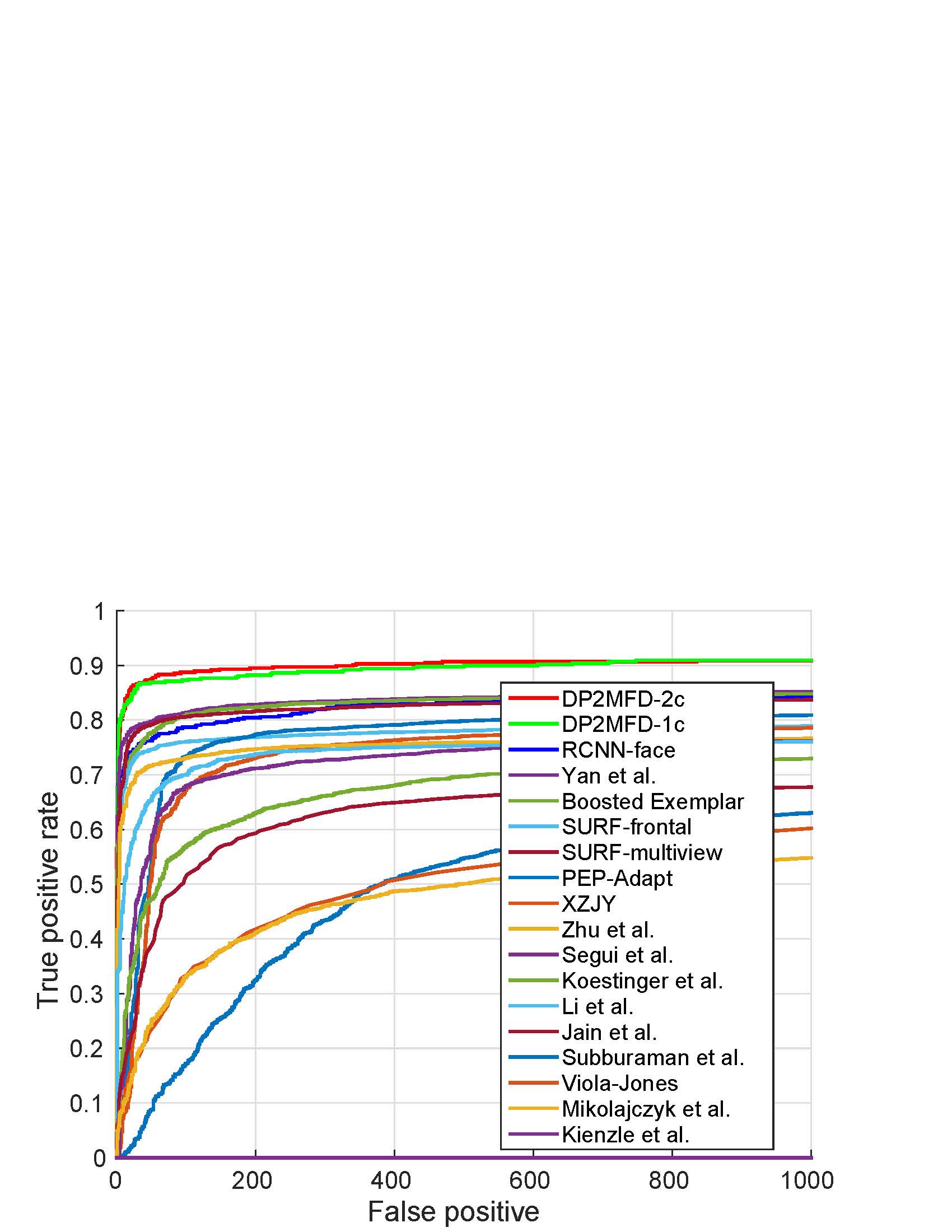}\hskip.1pt\includegraphics[width=4.2cm]{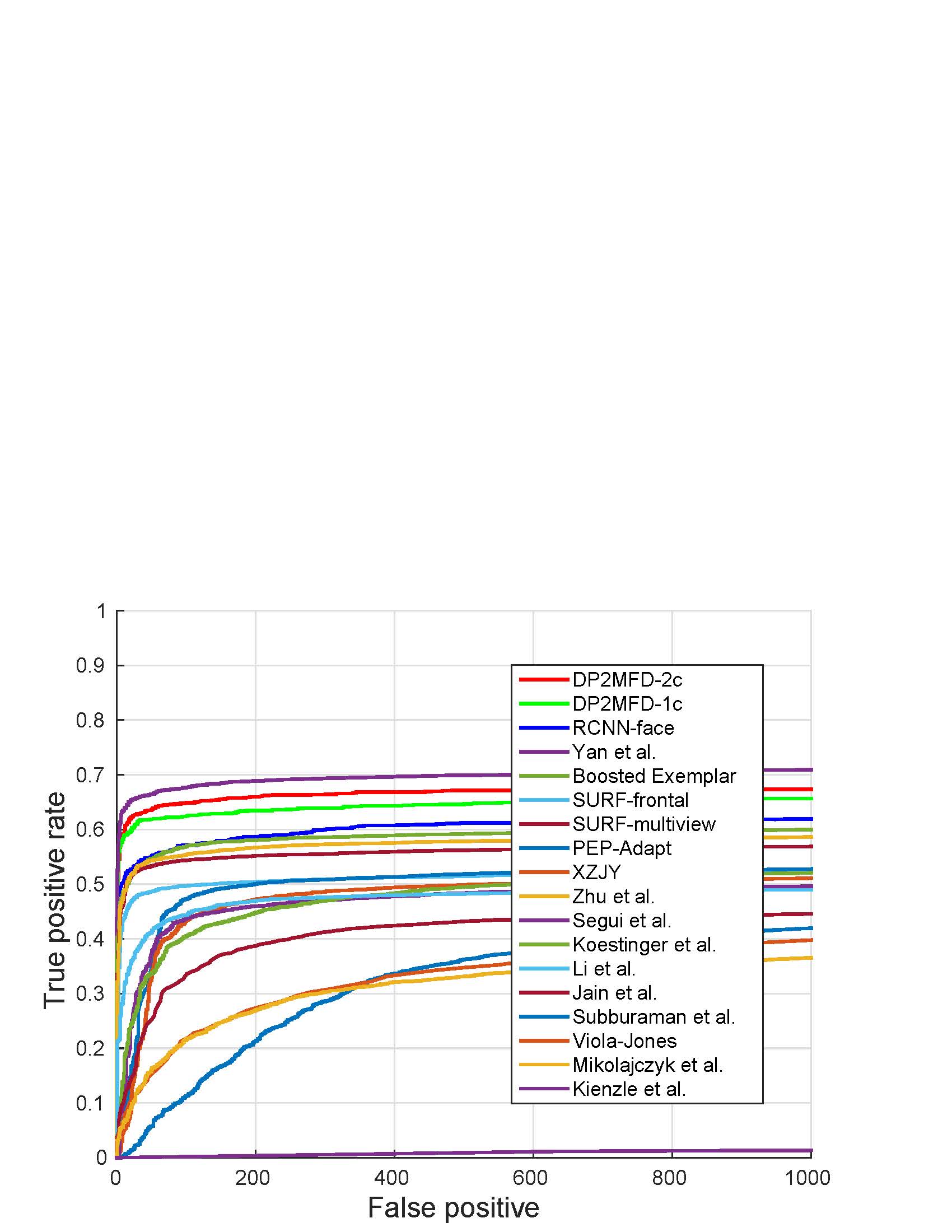}\hskip1pt\includegraphics[width=4.2cm]{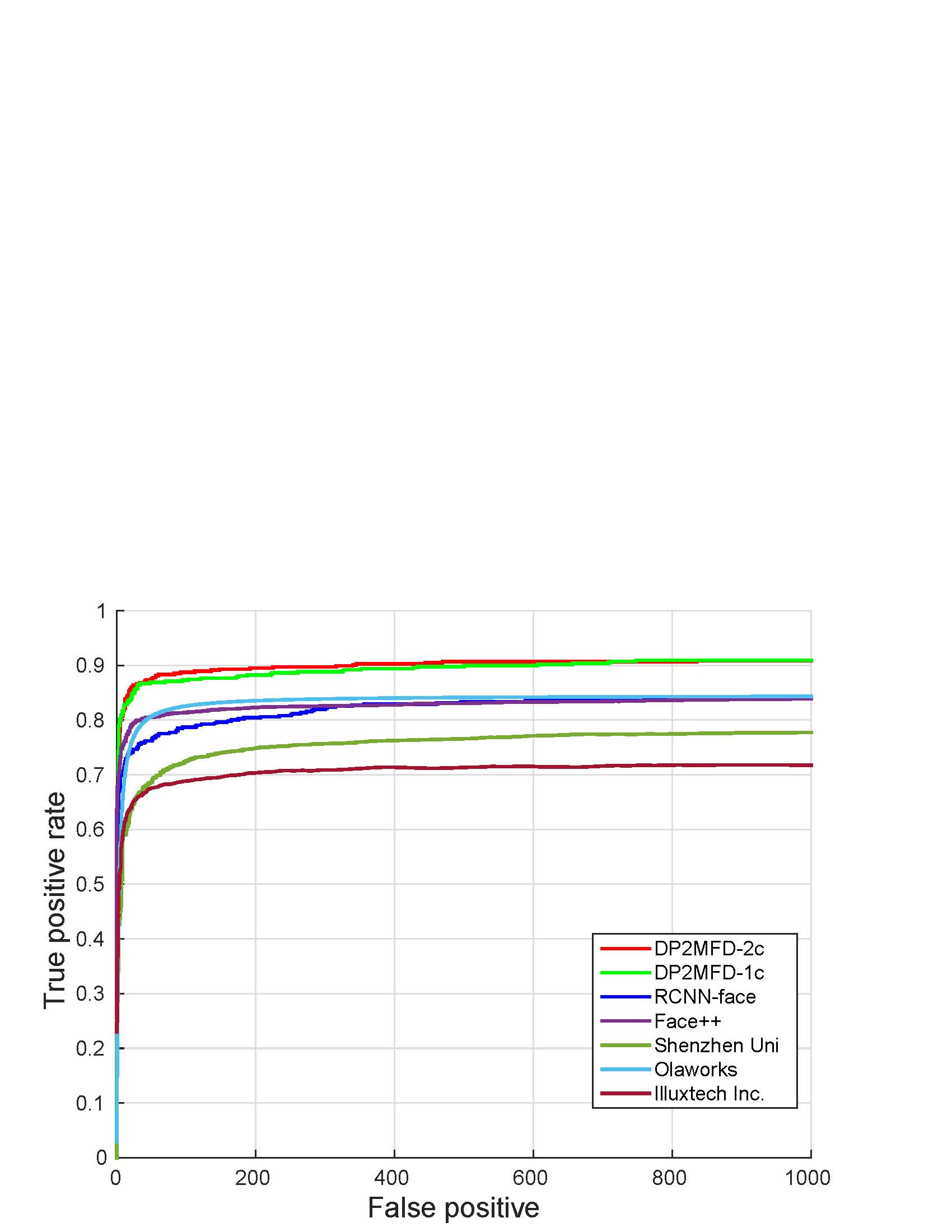}\hskip1pt\includegraphics[width=4.2cm]{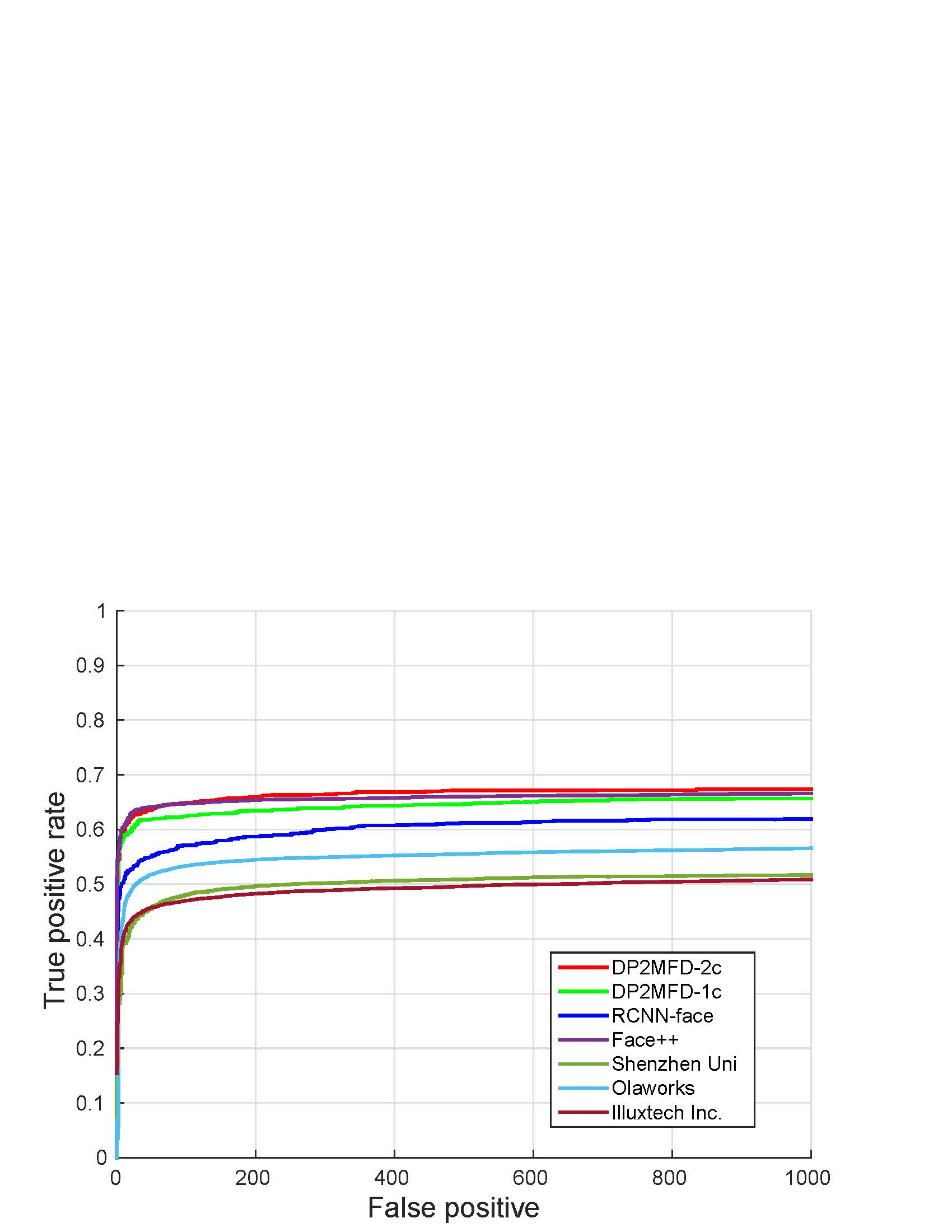}\\
(a)\hskip100pt(b)\hskip100pt(c)\hskip100pt(d)\\
\caption{Performance evaluation on the FDDB dataset.  (a) and (b) compare our method with previously published methods under the discrete and continuous protocols, respectively.  Similarly, (c) and (d) compare our method with commercial systems under the discrete and continuous protocols, respectively. }
\label{fig:FDDB_results}
\end{figure*}

Figure~\ref{fig:FDDB_results} compares the performance of different academic and commercial detectors using the Receiver Operating Characteristic (ROC) curves on this dataset.  The academic algorithms compared in Figure~\ref{fig:FDDB_results}(a)-(b) include Yan et al. \cite{Yan_CVPR2014}, boosted exemplar \cite{BoostedExemplar_Li_CVPR2014}, SURF frontal and multi-view \cite{SURF_frontal_multiview_Li_CVPR2013}, PEP adapt \cite{PEP_Adapt_LI_ICCV2013}, XZJY \cite{XZJY_Shen_CVPR2013}, Zhu et al. \cite{AFW_dataset_CVPR2012}, Segui et al. \cite{Segu'_ICPRAM2012}, Koestinger et al. \cite{Kšstinger_robustface_DAGM2012}, Li et al. \cite{LI_ICCV_workshop2011}, Jain et al. \cite{Jain_CVPR2011}, Subburaman et al. \cite{ subburaman_eccv_workshop_2010}, Viola-Jones \cite{Viola_Jones}, Mikolajczyk et al.  \cite{Mikolajczyk_ECCV2004}, Kienzle et al.  \cite{Kienzle_NIPS2005}  and the commercial algorithms compared in Figure~\ref{fig:FDDB_results}(c)-(d) include Face++, the Olaworks face detector, the IlluxTech frontal face detector and the Shenzhen University face detector \footnote{http://vis-www.cs.umass.edu/fddb/results.html}.

As can be seen from this figure, our method significantly outperforms all previous academic and commercial detectors under the discrete protocol and performs comparably to the previous state-of-the-art detectors under the continuous protocol. A decrease in performance for the continuous case is mainly because of low IOU score obtained in matching our detectors' rectangular bounding box with elliptical ground truth mask for the FDDB dataset.

We also implemented an R-CNN method for face detection and evaluated it on the FDDB dataset. The R-CNN method basically selects face independent candidate regions from the input image and computes a 4096 dimensional $fc_{7}$ feature vector for each of them. An SVM trained on $fc_{7}$ features classifies each region as face or non-face based on the detection score. The method represented by ``RCNN-face" performs better than most of the academic face detectors \cite{AFW_dataset_CVPR2012,SURF_frontal_multiview_Li_CVPR2013,PEP_Adapt_LI_ICCV2013}. This shows the dominance of deep CNN features over HOG, SURF. However, RCNN-Face's performance is inferior to the DP2MFD method as the region selection process might miss a face from the image.

\subsection{MALF Dataset Results}
The MALF dataset  \cite{MALF}  consists of 5,250 high-resolution images containing a total of 11,931 faces.  The images were collected from Flickr and image search service provided by Baidu Inc.   The average image size in this dataset is $573\times638$.  On average, each image contains 2.27 faces with $46.97\%$ of the images contain one face, $43.41\%$ contain 2 to 4 faces, $8.30\%$ contain 5 to 9 faces and $1.31\%$ images contain more than 10 faces.  Since this dataset comes with multiple annotated facial attributes, evaluations on attribute-specific subsets are proposed.   Different subsets are defined corresponding to different  combinations of attribute labels.   In particular,  `easy' subset contains faces without any large pose, occluded or exaggerated expression variations and are larger than $60\times60$ in size and `hard' subset contains faces that are larger than $60\times60$ in size with one of extreme pose or expression or occlusion variations.  Furthermore, scale-specific evaluations are also proposed in which algorithms are evaluated on two subsets - `small' and `large'.  The `small' subset contains images that have size smaller than $60\times60$ and the `'large' subset contains images that have size larger than $90\times90$.

\begin{figure*}[htp!]
 \centering
\includegraphics[width=5.5cm]{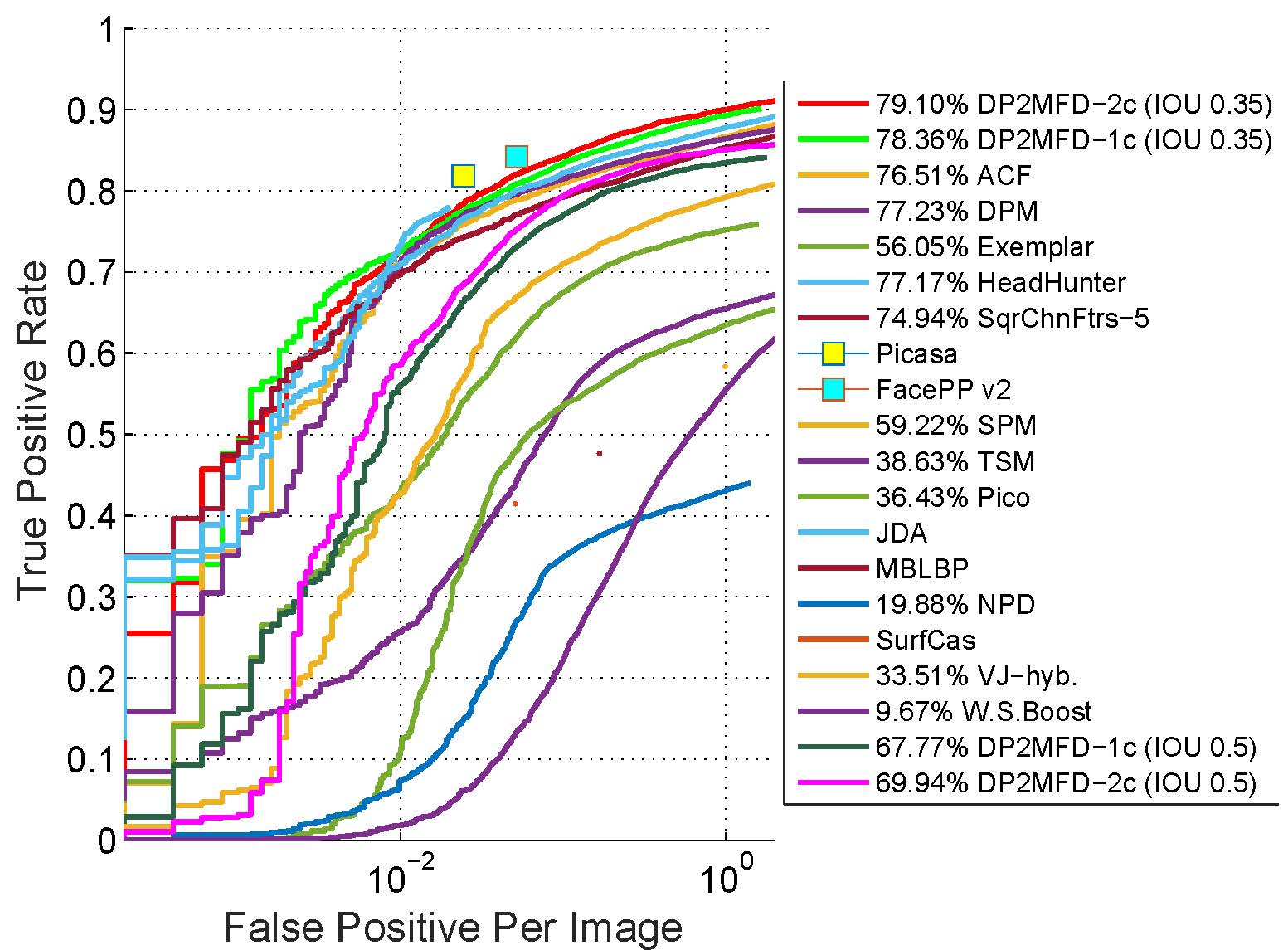}\hskip1pt\includegraphics[width=5.5cm]{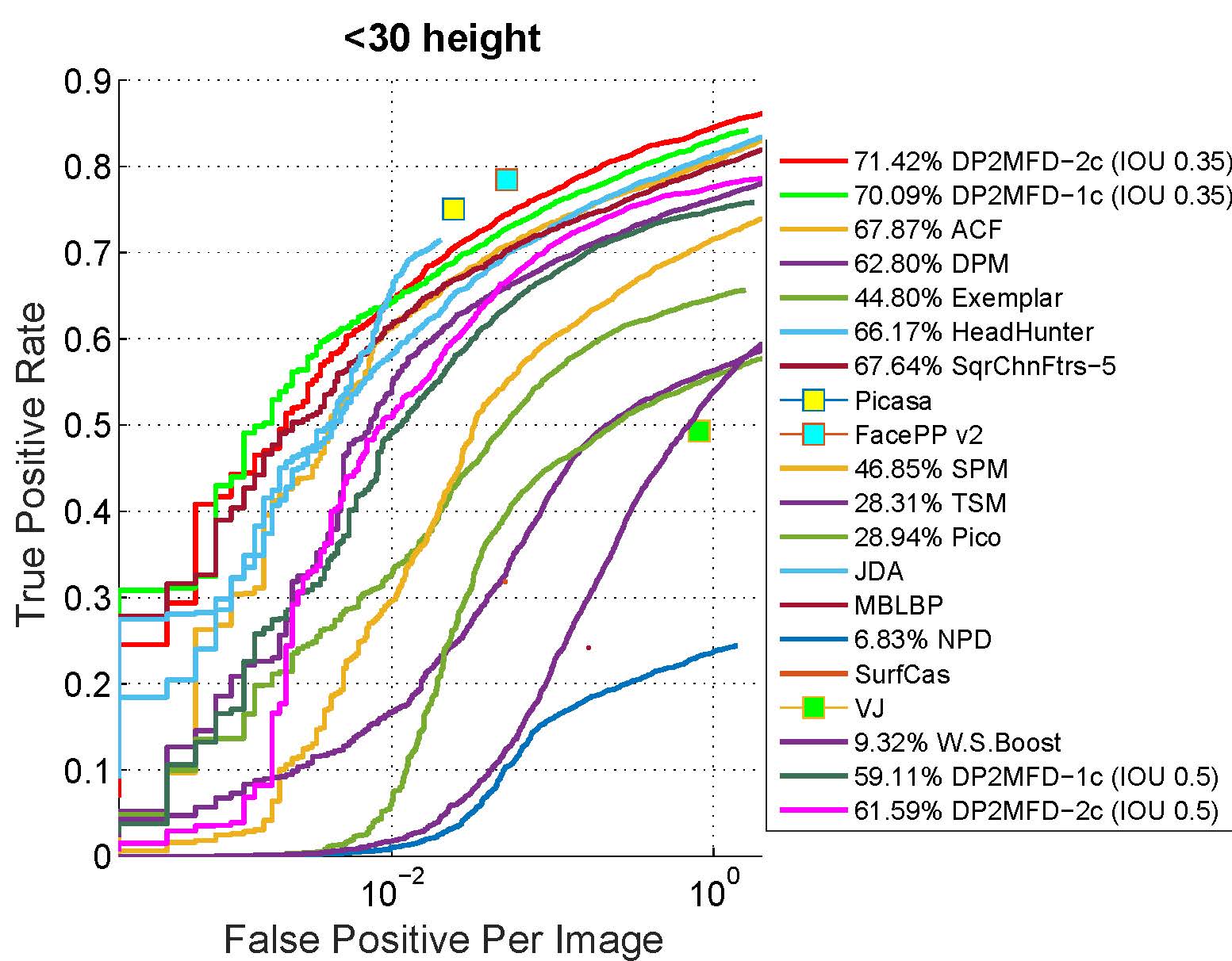}\\
(a)\hskip140pt(b)\\
\includegraphics[width=5.5cm]{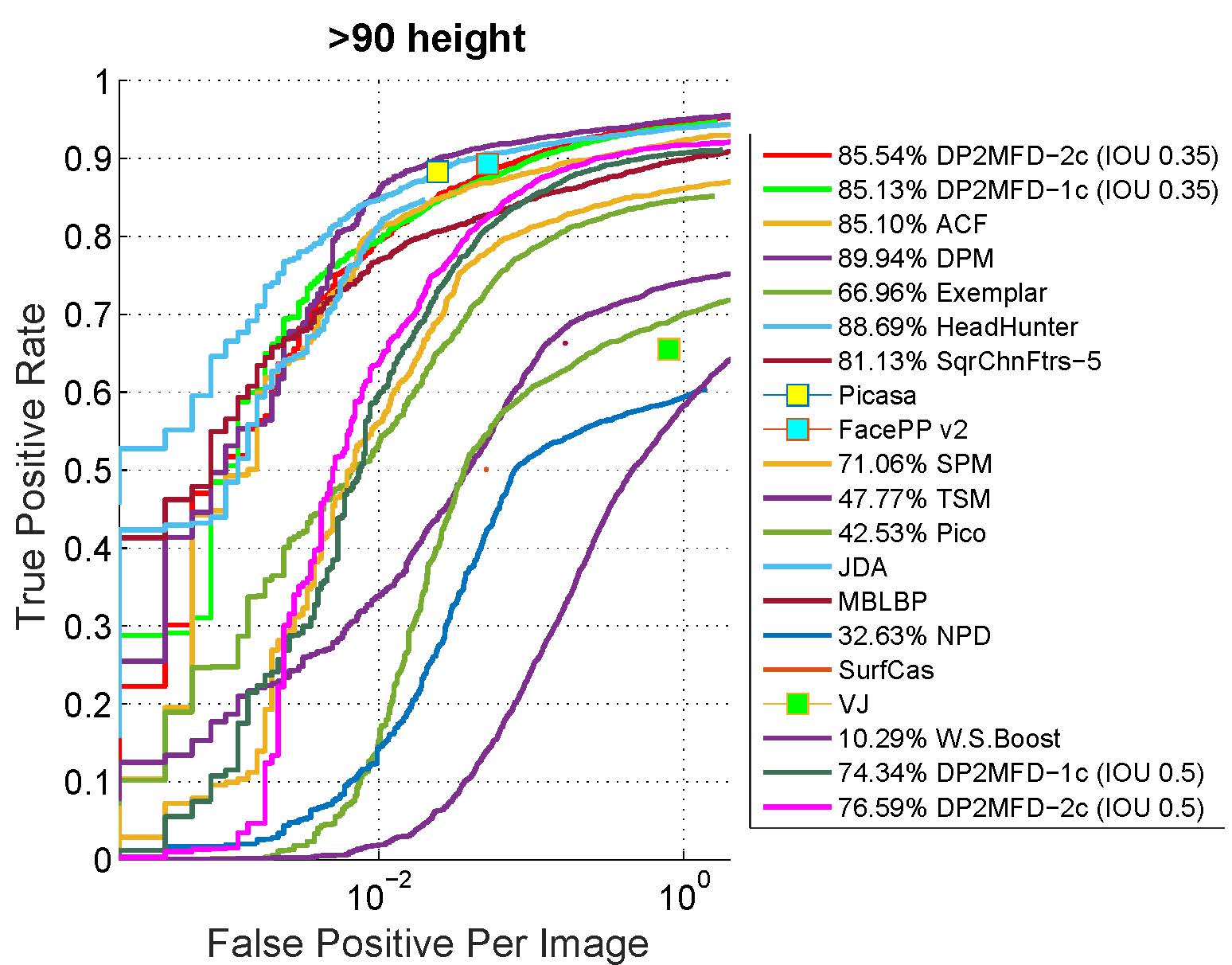}\hskip5pt\includegraphics[width=5.5cm]{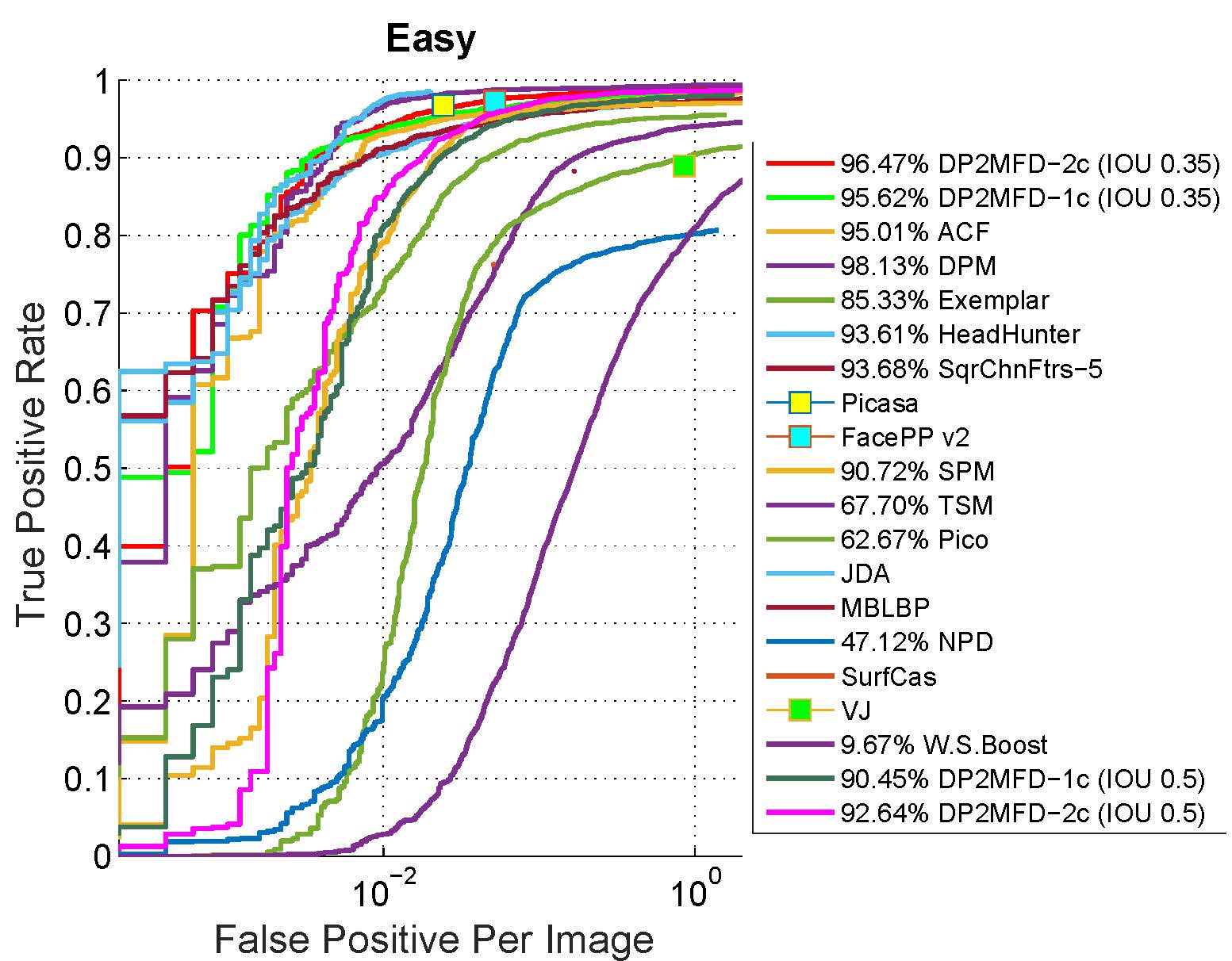}\hskip5pt\includegraphics[width=5.5cm]{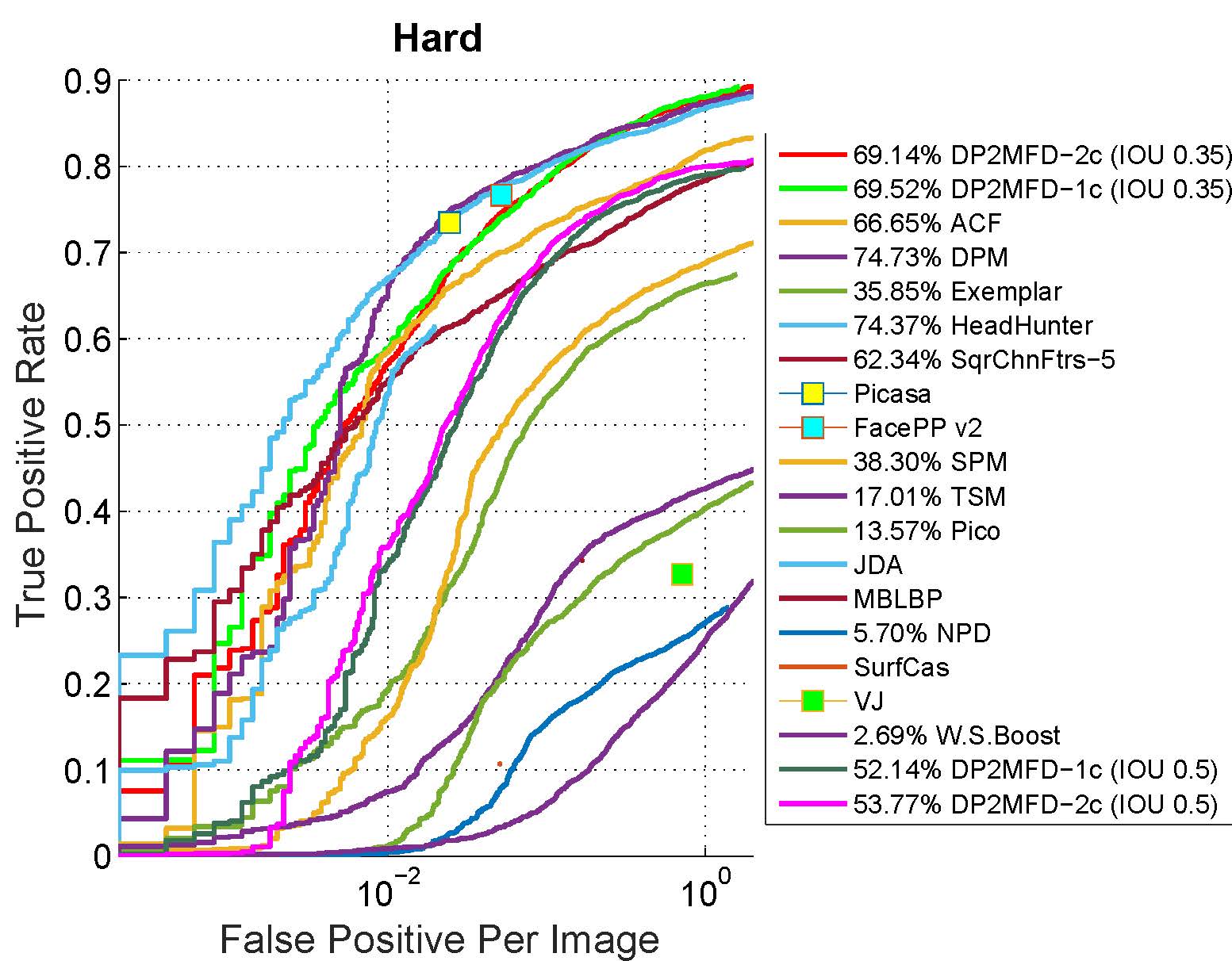}\\
(c)\hskip140pt(d)\hskip140pt(e)
\caption{Fine-grained performance evaluation on the MALF dataset.  (a) on the whole test set, (b) on the small faces sub-set, (c) on the large faces sub-set, (d)  on the `easy' faces sub-set and (e) on the `hard' faces sub-set.}
\label{fig:MALF_results}
\end{figure*}

The performance of different algorithms, both from academia and industry, are compared in Figure~
\ref{fig:MALF_results} by plotting the True Positive Rate vs. False Positive Per Images curves \footnote{The results of the methods other than our DP2MFD methods compared in Figure~\ref{fig:MALF_results} were provided by the authors of \cite{MALF}.}.    Some of the academic methods compared in Figure~\ref{fig:MALF_results} include ACF \cite{ACF_multiscale_Yang_IJCB2014}, DPM \cite{HeadHunter_Mathias_ECCV2014}, Exemplar method \cite{BoostedExemplar_Li_CVPR2014}, Headhunter \cite{HeadHunter_Mathias_ECCV2014}, TSM \cite{AFW_dataset_CVPR2012}, Pico \cite{PICO_Markus_CoRR2014},  NPD \cite{NPD2015} and W. S. Boost \cite{Kalal_BMVC2008}.
From Figure~\ref{fig:MALF_results}(a), we see that overall the performance of our DP2MFD method is the best among the academic algorithms and is comparable to the best commercial algorithms FacePP-v2 and Picasa.

In the `small' subset, denoted by $<30$ height in Figure~\ref{fig:MALF_results}(b), the performance of all algorithms drop a little but our DP2MFD method still performs the best among the other academic methods.  On the 'large', 'easy, and 'hard' subsets, the DPM method \cite{HeadHunter_Mathias_ECCV2014} performs the best and our DP2MFD method performs the second best as shown in Figure~\ref{fig:MALF_results}(c), (d) and (e), respectively. The DPM and Headhunter \cite{HeadHunter_Mathias_ECCV2014} are better as they train multiple models to fully capture faces in all orientations, apart from training on more than 20,000 samples.

We provide the results of our method for the IOU of $0.35$ as well as $0.5$ in Figure~\ref{fig:MALF_results}. Since the non-maximum suppression ensures that no two detections can have IOU$>0.3$, the decrease in performance for IOU of $0.5$ is mainly due to improper bounding box localization. One of the contributing factors might be the localization limitation of CNNs due to high amount of sub-sampling. In future, we plan to analyze this issue in detail.

\subsection{IJB-A Dataset Results}
The IJB-A dataset contains images and videos from 500 subjects collected from online media \cite{JanusFR_CVPR2015}, \cite{JanusFD_ICB2015}.   In total, there are 67,183 faces of which 13,741 are from images and the remaining are from videos.  The locations of all faces in the IJB-A dataset were manually ground truthed by human annotators.  The subjects were captured so that the dataset contains wide geographic distribution.   All face bounding boxes are about 36 pixels or larger.

\begin{figure*}[htp!]
 \centering
\includegraphics[width=6cm]{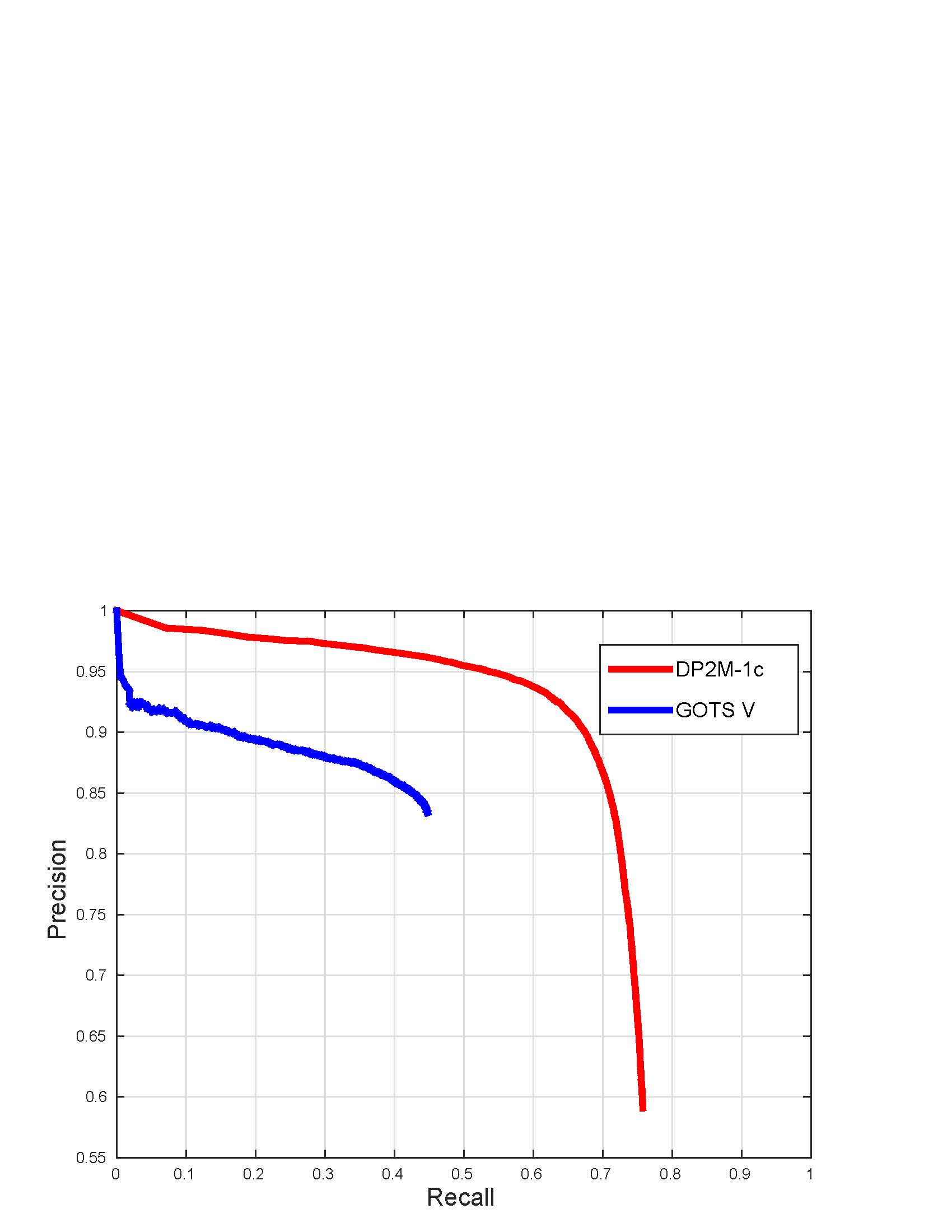}\hskip5pt\includegraphics[width=6cm]{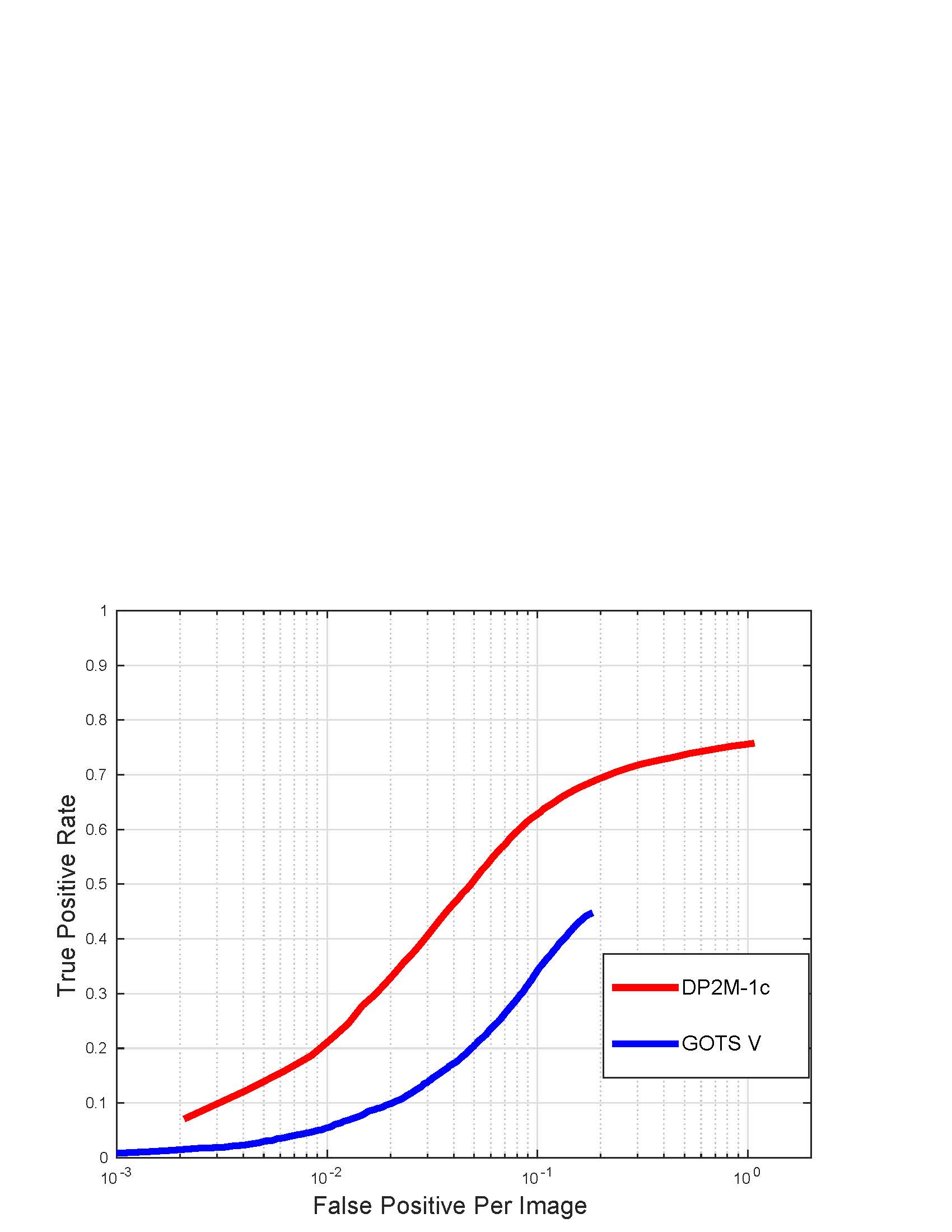}\\
(c)\hskip150pt(d)
\caption{Performance evaluation on the IJB-A dataset.  (a) Precision vs. recall curves. (b) ROC curves. }
\label{fig:IJBA_results}
\end{figure*}

Nine different face detection algorithms were evaluated on this dataset in \cite{JanusFD_ICB2015}.  Some of the  algorithms compared in \cite{JanusFD_ICB2015} include one commercial off the shelf (COTS) algorithm, three government off the shelf (GOTS) algorithms, two open source face detection algorithms (OpenCV's Viola Jones and the detector provided in the Dlib library), and PittPat ver 4 and 5.  In Figure~\ref{fig:IJBA_results} (a) and (b) we show the prevision vs. recall curves and the ROC curves, respectively corresponding to our method and one of the best reported methods in \cite{JanusFD_ICB2015}.   As can be seen from this figure, our method outperforms the best performing method reported in \cite{JanusFD_ICB2015} by a large margin.

\subsection{Discussion}

\begin{figure*}[htp!]
 \centering
 \includegraphics[width=16cm]{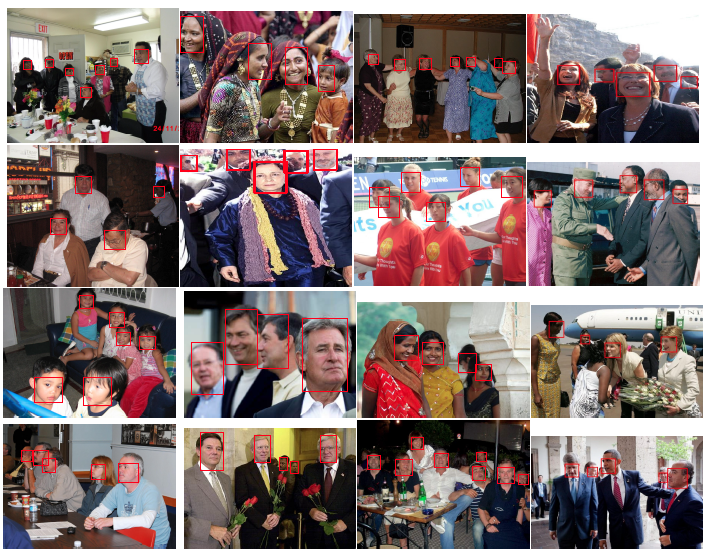}\\
\caption{Qualitative results of our detector.  First column - AFW dataset, Second column - FDDB dataset, Third column - MALF dataset, Fourth column - IJB-A dataset.}
\label{fig:qualitative}
\end{figure*}

Its clear from these results that our DP2MFD-2c method performs slightly better than the DP2MFD-1c method. This can be attributed to the fact that the aspect ratio of face doesn't change much with pose.  Figure~\ref{fig:qualitative} shows several detection results on the four datasets.  It can be seen from this figure, that our method is able to detect profile faces as well as different size faces in images with cluttered background.

\subsection{Runtime}
Our face detector was tested on a machine with 4 cores, 12GB RAM, and 1.6GHz processing speed. No GPU was used for processing. The model DP2MFD-1c took about 24.5s on average to evaluate a face, whereas DP2MFD-2c took about 26s. The deep pyramid feature evaluation took around 23s. It can certainly be reduced to 0.5s \cite{DPM_r_CNN} by using Tesla K20 GPU for feature extraction.

\section{Conclusions} \label{conclusion}
In this paper, we presented a method for unconstrained face detection which essentially trains DPM for faces on deep feature pyramid.   One of the interesting features of our algorithm is that we add a normalization layer to the deep CNN which reduces the bias in face sizes.  Extensive experiments on four publicly available unconstrained face detection datasets demonstrate the effectiveness of our proposed approach.

Our future work will include a GPU implementation of our method for reducing the computing time.   We will also evaluate the performance of our method on other object detection datasets.

\section*{Acknowledgments}
This research is based upon work supported by the Office of the Director of National Intelligence (ODNI), Intelligence Advanced Research Projects
Activity (IARPA), via IARPA R\&D Contract No. 2014-14071600012. The views and conclusions contained herein are those of the authors and should
not be interpreted as necessarily representing the official policies or endorsements, either expressed or implied, of the ODNI, IARPA, or the U.S. Government. The U.S. Government is authorized to reproduce and distribute reprints for Governmental purposes notwithstanding any copyright annotation
thereon.

{\small
\bibliographystyle{ieee}
\bibliography{FD}
}

\end{document}